# From Self-ception to Image Self-ception: A method to represent an image with its own approximations

Hamed Shah-Hosseini

tasom2002@yahoo.com, shahhosseini@srbiau.ac.ir

www.linkedin.com/in/dr-hamed-shah-hosseini

**Abstract**- *A concept of defining images based on its own approximate ones is proposed here, which is called "Self-ception". In this regard, an algorithm is proposed to implement the self-ception for images, which we call it "Image Self-ception" since we use it for images. We can control the accuracy of this self-ception representation by deciding how many segments or regions we want to use for the representation. Some self-ception images are included in the paper. The video versions of the proposed image self-ception algorithm in action are shown in a youtube channel* (find it by Googling image self-ception).

## 1. Introduction

The self-ception is proposed here as a concept by which any object or entity can be represented by its own approximate ones. It is like defining a function with its own approximate versions of the original function. So, conceptually, if we have a function, like $f(x)$; then, we may represent it by the following summation :

$$f(x) \approx T_1\big(f(x)\big) + T_2\big(f(x)\big) + \cdots . T_N(f(x))$$

(1)

Where $T_i$ are transformations that are applied to the given function.

The question that remains is that what the transformations $T_i$ should be, or how we can find or define them. Moreover, the number of these transformations, denoted by $N$ in Eq. (1) is user-selected.

In this paper, we propose an algorithm to implicitly define the transformations $T_i$. Thus, the whole function, which is an image here, is represented by these components. Specifically, an image self-ception algorithm is proposed here. Next section reviews related work. Section 3 introduces the steps of the proposed image self-ception algorithm. Experiments are included in section 4. Finally, section 5 contains the concluding remarks.

## 2. Related work

The concept of self-ception may resemble that of self-similarity in Fractal Geometry [1]. However, the theories behind Fractal Geometry are not flexible enough to enable us to reach the self-ception and to develop an algorithm for directly implementing the self-ception proposed here.

It should be mentioned a method called "imageception" is introduced in [2], in which an image is represented with other images. But,

here, we emphasize to use the same image as components to represent the whole original image.

Next section, section 3, expresses the proposed image self-ception algorithm.

## 3. The Proposed method: Image Self-ception

To define an image based on its own approximated ones, we may execute the following steps. However, this is just one way to do the self-ception, and researches may use other ways to implement the concept of image self-ception.

The proposed method, Image self-ception, may be expressed in the following steps:

1) Divide the image into several segments. An image segmentation algorithm should be used to do the segmentation. However, depending on the parameters used in the image segmentation algorithm, the number of segments of the original image may be different. Therefore, the number of segments are somehow user-defined, either explicitly or implicitly.

2) Approximate each segment found by step (1), by an ellipse. This can be done by finding the principal axis of the segment and its orientation.

It is mentioned that approximating with ellipses is just one option among many options.

3) Find the bounding box of each ellipse of a segment. Then, resize the original image into this bounding box. At the end of this step, for each ellipse, we have a resized image of the original image.

However, one could find the smallest bounding box of each ellipse. Then, rotate the image by the orientation of ellipse; then, resize the image. But, for now, we use the former and simpler approach.

4) Here, we use an approximation to color updating. The pixels inside each region of an ellipse has a mean color. In addition, the original image has also a mean color. We can simply add each color of resized image with the vector difference between two means. Specifically:

$$color_{ellipse}(x,y) \leftarrow color_{resized\ image}(x,y) + (rang_{ellipse} - Total_{mean\ color})$$

$$\forall x, y \in region\ of\ ellipse \quad (2)$$

Where $rang_{ellipse}$ is the 3D mean color of pixels inside ellipse. Also, $Total_{mean\ color}$ is the 3D mean color of the original image. Moreover, $(x,y)$ denotes the position of each pixel, and $color_{ellipse}(x,y)$ denotes the 3D color value of that pixel inside the ellipse whereas $color_{resized\ image}(x,y)$ is the 3D color of the resized image at point $(x,y)$.

Since the resized image is a small version of the original image, its total mean color ought to be close to that of the original image. That is why we have used $Total_{mean\ color}$ in equation above as an approximation to the mean color of each resized image. On may conclude that Eq. (2) makes the mean color of the ellipse almost equal to $rang_{ellipse}$.

It is reminded that we could use Histogram Matching [3] instead of Eq. (2), but it takes more computational power.

Next section, section 4, includes images produced by the proposed image self-ception algorithm.

## 4. Experiments

In this section, we represent a few images that are generated by the proposed algorithm.

For image segmentation of step (1) of the proposed algorithm, we use the method SLIC, which is a K-means based image segmentation algorithm [4]. This method has an implementation in Scikit-image [5].

For step (2) of the proposed image self-ception, we need to find the best ellipses that cover segments (or regions) produced from the image segmentation step. The center of each ellipse is the region's centroid, and its orientation is defined by the direction of the major axis. In summary, we can find the best ellipse here by using the eigenvalues of a matrix formed by central moments of the given region. The procedure to find the best ellipse is mentioned in several Image processing and Computer Vision textbooks including in [6].

The steps (3) and (4) have been mentioned in previous section, and are straight-forward to implement.

First experiment is shown in Figures 1 to 4 in which an image of a cat, named Chelsea of Scikit-image, is undergone self-ception.

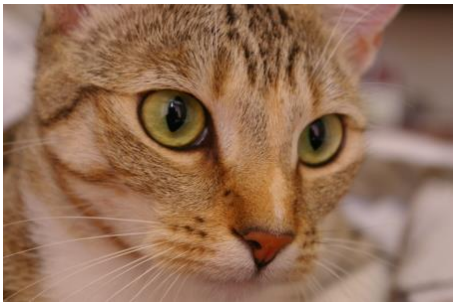

Figure 1. The cat named Chelsea.

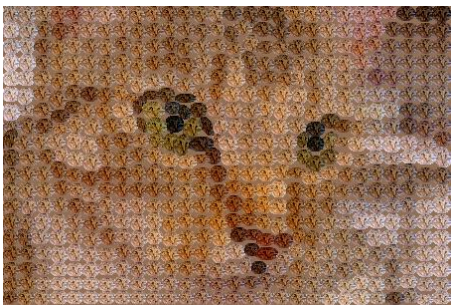

Figure 2. Self-ception of Figure 1 with 532 regions. Here, every region is a small and/or rotated version of the original image of Chelsea. The MSE of this image with Figure 1 is 1235.97.

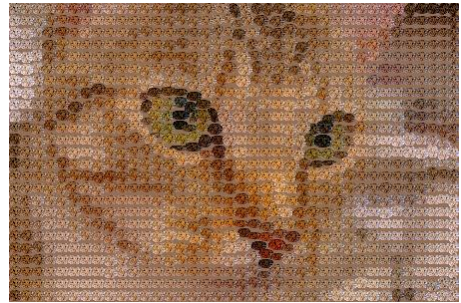

Figure 3. Self-ception of Figure 1 with 950 regions. The MSE of this image with Figure 1 is 1111.39.

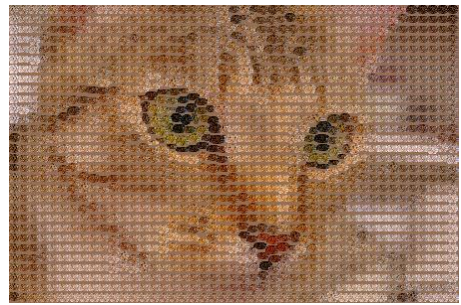

Figure 4. Self-ception of Figure 1 with 1349 regions. The MSE between this image and Figure 1 is 1044.12.

Based on the MSEs (Mean-Squared Errors) reported in figure captions, in Figures from 2 to 4, as the number of regions increases, the self-ception image becomes closer to the original image, but visually we lose the details of each region itself.

A similar experiment is performed on Figure 5, the Coffee image of Scikit-image, to get self-ception images of Figures 6 to 8. And as the number of regions increases, the MSE values also decrease.

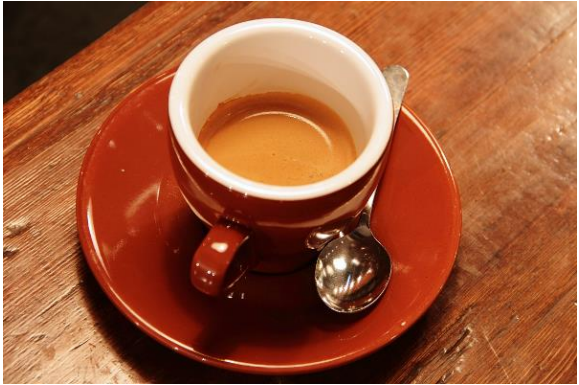

Figure 5. The original Coffee image.

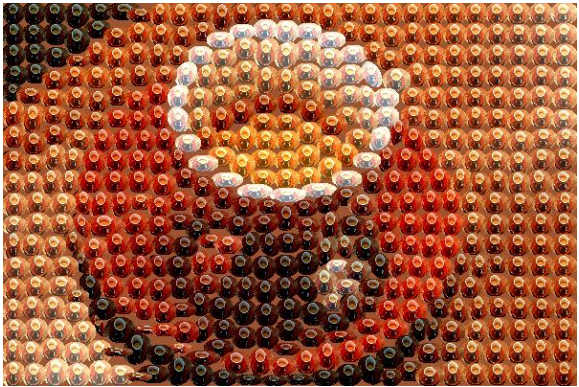

Figure 6. The self-ception image of Figure 5 produced by 485 regions and MSE 3489.76.

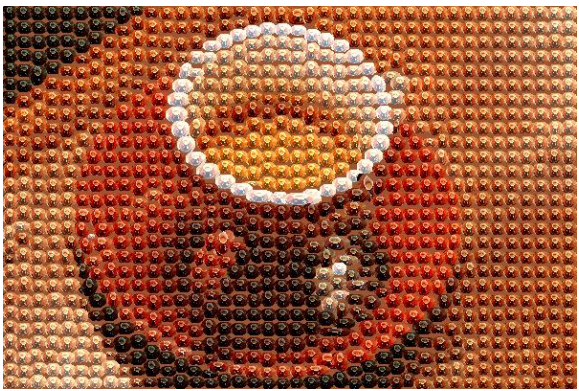

Figure 7. The self-ception image of Figure 7 with 1057 regions and MSE 3321.85.

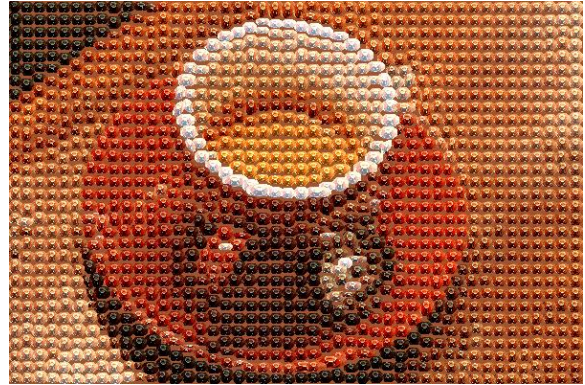

Figure 8. The self-ception image of Figure 5 with 1406 regions and MSE 3262.42.

## 5. Conclusion

A new concept was defined here called "Self-ception". When we use this concept for images, we call it "Image Self-ception". The steps to implement it were included in the paper. Experiments showed that an image can indeed be represented by its own approximate ones. The accuracy depends on how many components we are desired to use. So, it is a user-selected option.

The self-ception proposed here should not be limited to only images. In fact, researchers may use it for other kinds of media, including sounds and videos. Or even, one day we may be able to use it for physical concepts and physical objects.

## References

[1] K. Falconer. Fractal Geometry: Mathematical Foundations and Applications. John Wiley & Sons, 2003.

[2] https://github.com/mattyhall/Imageception

[3] R.C. Gonzalez, and R.E. Woods, Digital Image Processing. 3rd ed. Upper Saddle River, N.J.: Prentice Hall, 2008.

[4] R. Achanta, A. Shaji, K. Smith, A. Lucchi, P. Fua, and S. Süsstrunk. "SLIC Superpixels

Compared to State-of-the-art Superpixel Methods", TPAMI, May 2012.

[5] Scikit-image reference at: http://scikit-image.org/

[6] W. Burger and M. Burge. Principles of Digital Image Processing: Core Algorithms. Springer-Verlag, London, 2009.

# Appendix A

A figure of tile painting of a wall of the Golestan Palace in Tehran is shown in Figure 9. Then, a self-ception image of that figure is shown in Figure 10.

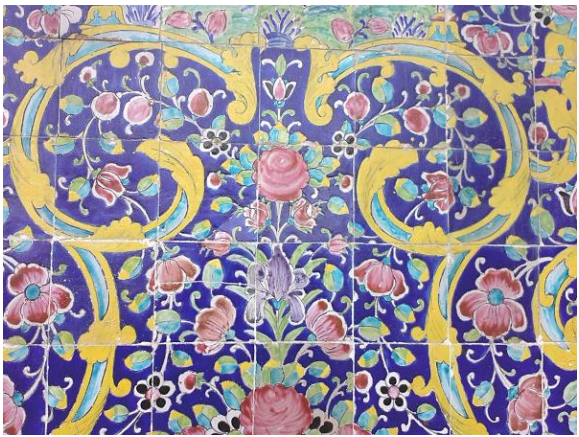

Figure 9. The original image of a wall of the Golestan palace in Tehran.

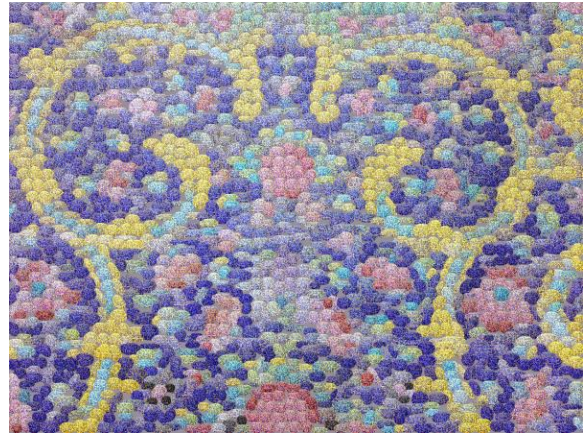

Figure 10. A self-ception image of Figure 9 with 1889 regions.